\documentclass[a4paper, 10pt, conference]{IEEEtran}
\IEEEoverridecommandlockouts
\usepackage{cite}
\usepackage{amsmath,amssymb,amsfonts}
\usepackage{algorithmic}
\usepackage{graphicx}
\usepackage{textcomp}
\usepackage{tabularx}
\usepackage{amsmath}

\usepackage{multirow}
\usepackage{dblfloatfix}
\usepackage{siunitx}
\usepackage{booktabs}
\usepackage{longtable}
\usepackage{cite}
\usepackage{color,colortbl}
\definecolor{Gray2}{gray}{0.9}
\definecolor{Gray}{gray}{0.7}

\usepackage{xcolor}
\def\BibTeX{{\rm B\kern-.05em{\sc i\kern-.025em b}\kern-.08em
    T\kern-.1667em\lower.7ex\hbox{E}\kern-.125emX}}
\begin{document}

\title{DandelionTouch: High Fidelity Haptic Rendering of Soft Objects in VR by a Swarm of Drones
}

\makeatletter
\newcommand{\linebreakand}{%
  \end{@IEEEauthorhalign}
  \hfill\mbox{}\par
  \mbox{}\hfill\begin{@IEEEauthorhalign}
}
\makeatother

\author{

    \IEEEauthorblockN{Aleksey Fedoseev}
    \IEEEauthorblockA{\textit{Intelligent Space Robotics Laboratory} \\
    \textit{Skoltech}\\
    Moscow, Russian Federation \\
    aleksey.fedoseev@skoltech.ru}
\and
    \IEEEauthorblockN{Ahmed Baza}
    \IEEEauthorblockA{\textit{Intelligent Space Robotics Laboratory} \\
    \textit{Skoltech}\\
    Moscow, Russian Federation \\
    ahmed.baza@skoltech.ru}
\and
    \IEEEauthorblockN{Ayush Gupta}
    \IEEEauthorblockA{\textit{Intelligent Space Robotics Laboratory} \\
    \textit{Skoltech}\\
    Moscow, Russian Federation \\
    ayush.gupta@skoltech.ru}
\linebreakand
    \IEEEauthorblockN{Ekaterina Dorzhieva}
    \IEEEauthorblockA{\textit{Intelligent Space Robotics Laboratory} \\
    \textit{Skoltech}\\
    Moscow, Russian Federation \\
    ekaterina.dorzhieva@skoltech.ru}
\and  

    \IEEEauthorblockN{Riya Neelesh Gujarathi}
    \IEEEauthorblockA{\textit{Imperial College London} \\ 
    Exhibition Road, South Kensington\\ 
    London,United Kingdom\\
    riya.gujarathi18@imperial.ac.uk}
    
\and
    \IEEEauthorblockN{Dzmitry Tsetserukou}
    \IEEEauthorblockA{\textit{Intelligent Space Robotics Laboratory} \\
    \textit{Skoltech}\\
    Moscow, Russian Federation \\
    d.tsetserukou@skoltech.ru}
}
\maketitle

\begin{abstract}
To achieve high fidelity haptic rendering of soft objects in a high mobility virtual environment, we propose a novel haptic display DandelionTouch. The tactile actuators are delivered to the fingertips of the user by a swarm of drones. Users of DandelionTouch are capable of experiencing tactile feedback in a large space that is not limited by the device's working area. Importantly, they will not experience muscle fatigue during long interactions with virtual objects.
Hand tracking and swarm control algorithm allow guiding the swarm with hand motions and avoid collisions inside the formation.\par
Several topologies of impedance connection between swarm units were investigated in this research. The experiment, in which drones performed a point following task on a square trajectory in real-time, revealed that drones connected in a Star topology performed the trajectory with low mean positional error (RMSE decreased by 20.6\% in comparison with other impedance topologies and by 40.9\% in comparison with potential field-based swarm control). The achieved velocities of the drones in all formations with impedance behavior were 28\% higher than for the swarm controlled with the potential field algorithm.\par
Additionally, the perception of several vibrotactile patterns was evaluated in a user study with 7 participants. The study has shown that the proposed combination of temporal delay and frequency modulation allows users to successfully recognize the surface property and motion direction in VR  simultaneously (mean recognition rate of 70\%, maximum of 93\%). DandelionTouch suggests a new type of haptic feedback in VR systems where no hand-held or wearable interface is required.
\end{abstract}


\section{Introduction}
Over the past decade, the performance of wearable haptic devices has been extensively improved, allowing them to present the experience of physical contact with virtual surfaces through various types of interaction, e.g., interaction with kinesthetic, vibrotactile, or electrotactile feedback \cite{Yin_2019}. However, such devices have to be attached to the human body, which adds extra weight and interferes with the motion of their arms. Thus, the virtual experience results in being less realistic to the user. 

\begin{figure}[ht]
 \includegraphics[width=1\linewidth]{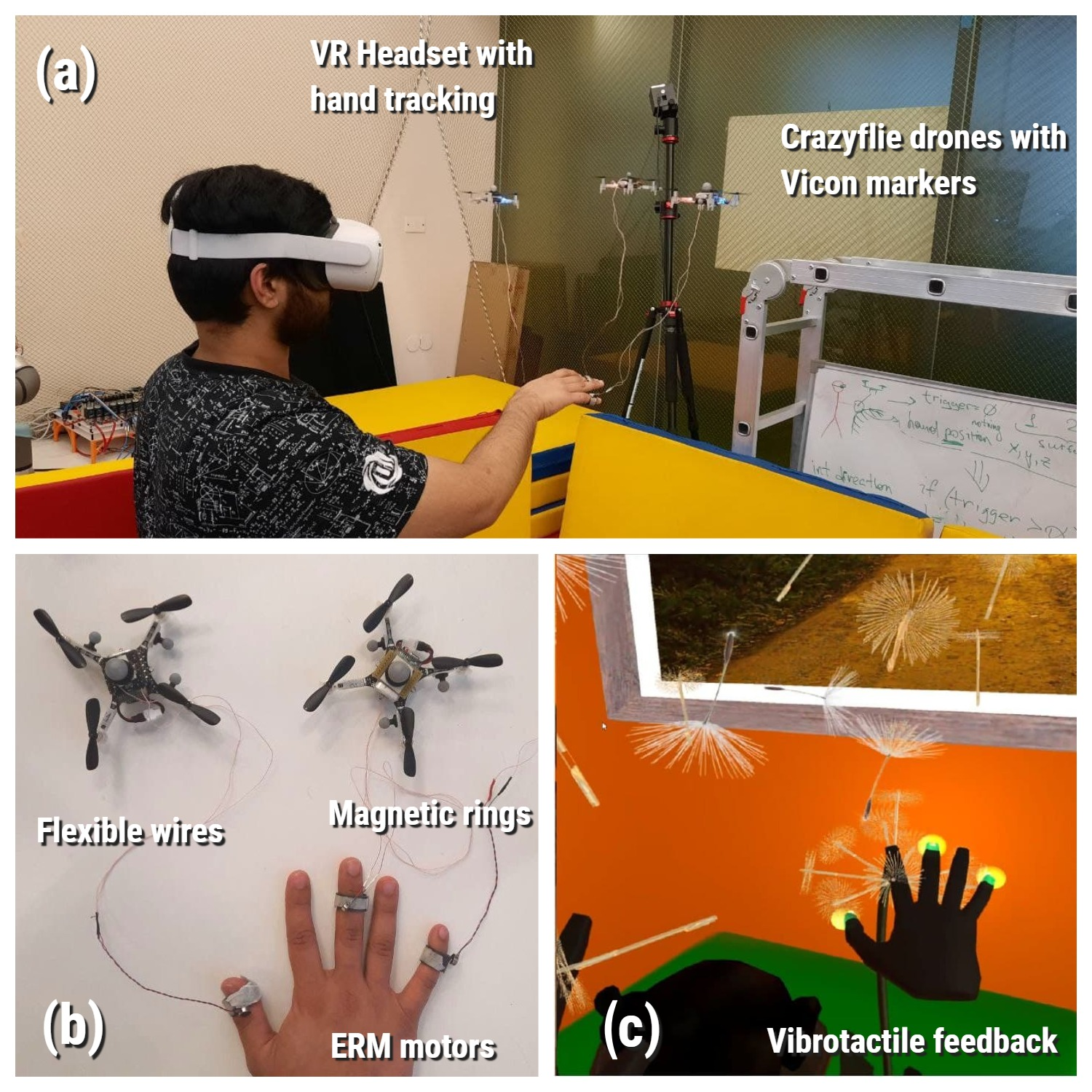}
 \caption{(a) The user receives tactile feedback when interacting with different agents in the VR environment. (b) Vibromotors are delivered to the user's finger by drones. (c) The user interacts with virtual objects through DandelionTouch. }
 \label{fig:ill5}
\end{figure}

Several researchers propose grounded \cite{Aggravi_2021}, encountered-type \cite{Furumoto_2019, Fedoseev_2019} and midair \cite{Takahashi_2020, Howard_2020} displays with high fidelity haptic feedback as an effective approach to distance haptic devices from users and to allow the users experience the freedom of motion for natural interaction with the environment. However, the working area of the mentioned above displays is still limited by their physical dimensions.

The recent research in human-drone interaction introduced drone-based systems such as BitDrones in \cite{Gomes_2016} and GridDrones in \cite{Braley_2018}, which can perform as input interfaces with high scalability in large working spaces. In virtual reality applications, such systems can perform as actively controlled displays for midair human-drone interactions similar to either encountered-type or exoskeleton devices. The scenario applications and the behavior of the drones in close proximity to the user are still a subject to be explored.

\section{Related Works}
Several concepts of haptic human-drone interaction (HDI) were suggested, such as the direct encounter of the user's hand with a drone explored by Abdullah et al. \cite{Abdullah_2018} for 1D gravitation force simulation. More complex systems apply additional haptic proxies attached to the drone for kinesthetic feedback, such as lightweight haptic extentions for passive and active feedback scenarios developed by Hoppe et al. \cite{Hoppe_2018}. Many applications for haptic drones were suggested by Yamaguchi et al. \cite{Yamaguchi_2016} in Virtual Reality (VR) scenario of interaction with a virtual sword and Abtahi et al. \cite{Abtahi_2019} in VR scenario of a virtual wardrobe. All the mentioned above scenarios allowed users to experience tactile interactions over a large area freely. However, they are focused on displays with limited contact forces supported by the rotors of drones. Additionally, the static form of such displays is not able to provide separate feedback to each of the user's fingers. Therefore, Tsykunov et al. \cite{Tsykunov_2019} proposed the concept of high-fidelity HDI-based display was proposed. However, the suggested kinesthetic feedback by wired drones remained sensitive to pulling forces by the human hand.\par
In this paper, we propose a novel concept of a haptic display in which drones deliver vibromotors on flexible cords to the user's fingertips. Unlike previous solutions, DandelionTouch eliminates the weight and bulk of wearable devices and achieves the robustness of HDI by providing high-fidelity vibrotactile feedback, thus lowering the force applied to the drones. The drone swarm follows the position of the user's hand with low latency, thereby delivering a wide range of vibration patterns upon contact with virtual surfaces.

\section{DandelionTouch Technology}

\subsection{System Overview}

The system architecture (Fig. \ref{fig:system}) is responsible for both providing the user with a real-time interactive VR environment and for controlling the Crazyflie drones. 

\begin{figure}[htb!]
 \includegraphics[width=\linewidth]{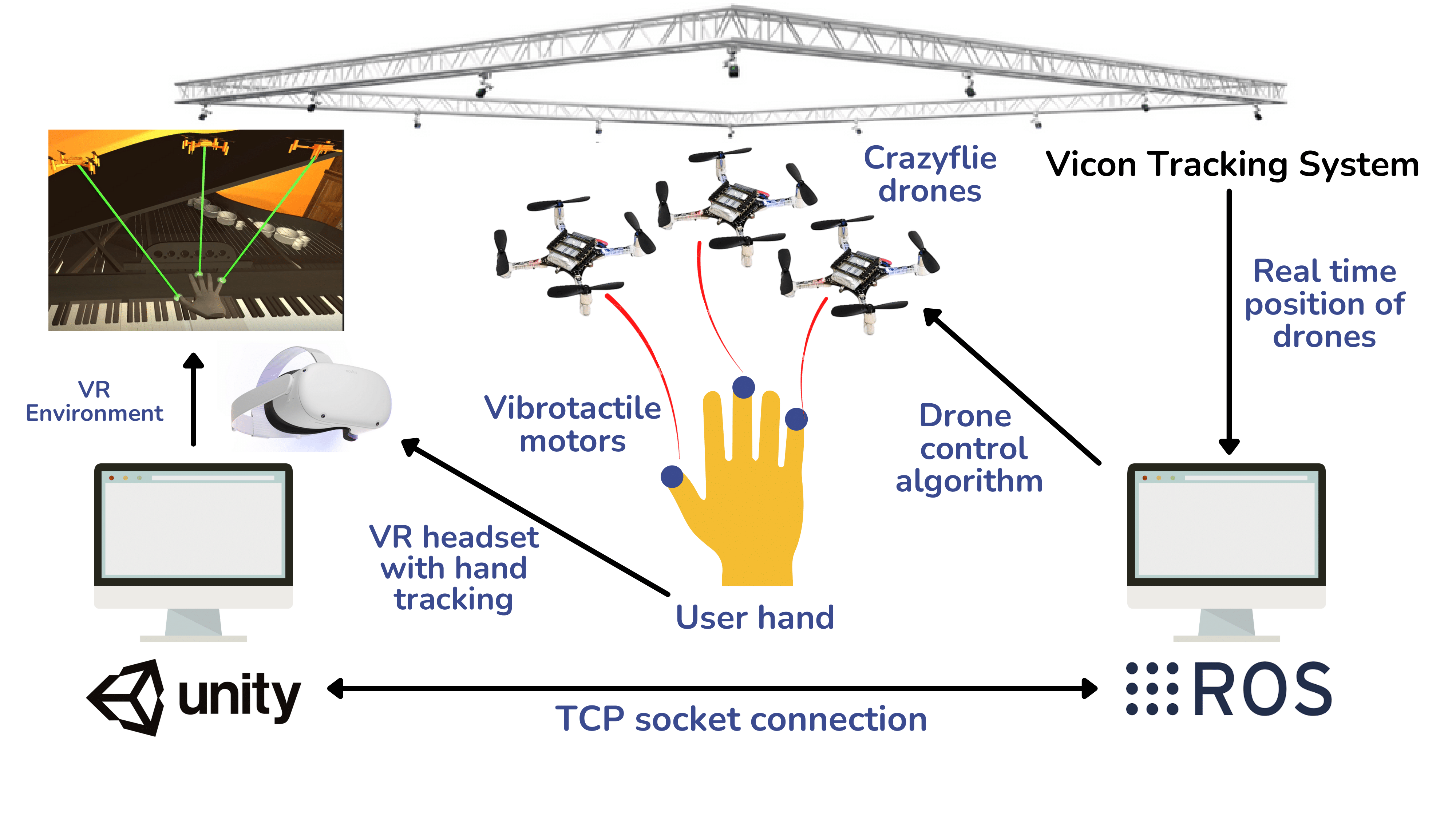}
 \caption{Overview of the DandelionTouch. User interacts with objects through VR framework with motion capture. Then ROS framework controls the swarm behavior to follow the user's hand and deliver feedback of interaction.}
 \label{fig:system}
\end{figure}

Unity game engine is used to develop and manage the VR environment that allows users to interact with virtual objects wearing the Oculus Quest or Vive Pro headsets with hand tracking. The connection between ROS and Unity is established through ROS-bridge and ROS-sharp libraries so that the Unity framework calculates the behavior of the swarm from the user motion and receives the real-time positions and orientations of the drones tracked by the VICON motion capture system.

Swarm behavior and vibrotactile feedback are handled with ROS subscribing to the real-time positions and orientations of the hand, the positions of the drones in the swarm, and the trigger messages from Unity. Upon a Unity trigger that the user is within a close distance to an interactive object, drones take off and attach the vibromotors to the user's fingertips. To avoid internal collision between swarm units and user's hand during motor attachment, the artificial potential field (APF) approach is applied. The swarm of drones then maintains the connection and follows the user's motions during the interaction using the impedance control model described in Section \ref{section:impedance}.

 \subsection{Haptic Rendering}
 \label{section:haptic_patterns}

The haptic rendering framework consists of two components: the VR environment and the drone swarm, connected via the TCP socket connection. The haptic rendering pipeline begins with generating an interactive VR environment in Unity, which includes several virtual objects with different surface properties. The pipeline initiates the collision detection and computes the motion direction of the user's hand, detected by the VR headset motion hand tracking. As the user enters the space of proximity interaction surrounding the virtual objects, the swarm automatically takes off and initiates hand tracking. When a collision is detected with any object in the VR environment, Unity identifies the object's surface type and, hence, its tactile properties. The hand motion direction and unique identifications of the surface are then sent to the drone swarm via the TCP socket to provide a haptic experience for the user. 

Each drone carries one ERM vibromotor on a flexible cord attached to a magnetic ring on the user's hand. Bitcraze Buzzer deck was modified by replacing the buzzer with vibromotor for generating different vibrotactile patterns instead of sounds to avoid additional weight on the drones. Modifying the buzzer deck and the Crazyflie firmware allowed us to display different vibration frequencies and control the vibromotors through the Crazyflie python API. Several approaches have been explored for surface simulation with vibrotactile actuators \cite{Basdogan_2020} to provide a sensation of stiffness in \cite{Maereg} or texture roughness in \cite{Asano_2015}. In this work, we propose to combine the temporal delay of the signal with frequency modulation to simultaneously provide the user with the experience of surface roughness and the direction of their hand motion. The vibration patterns for haptic feedback are presented in Fig. \ref{fig:haptic_ill}.

\begin{figure}[h!]
 \includegraphics[width=1\linewidth]{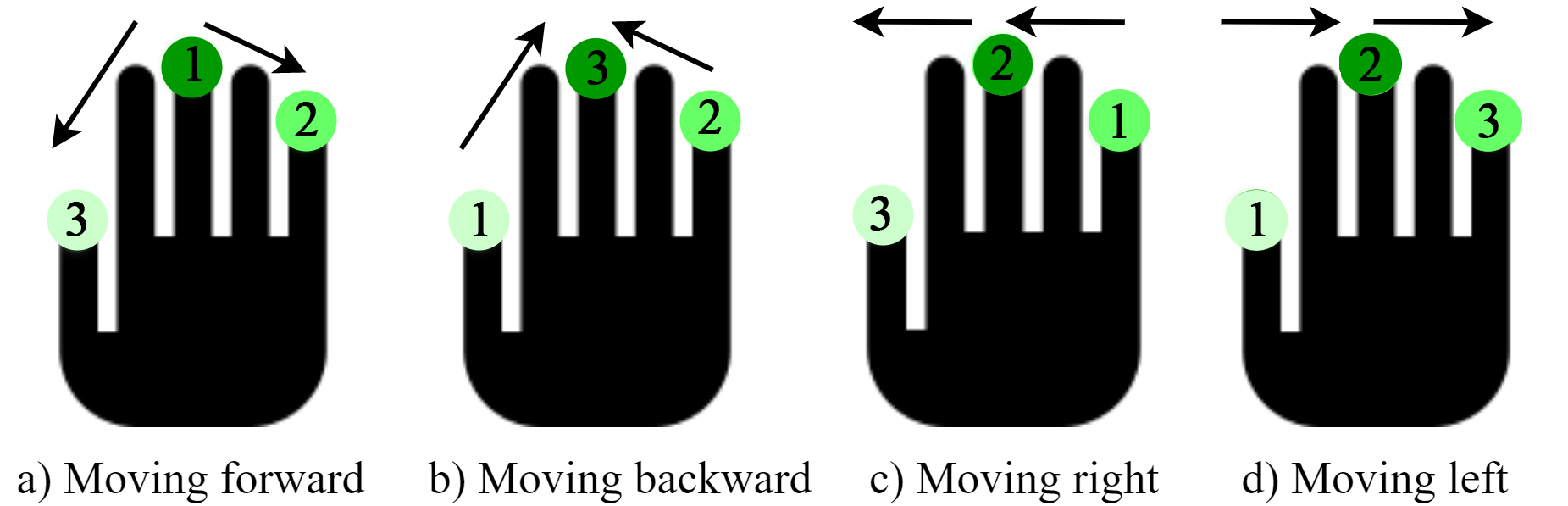}
 \caption{Tactile simulation patterns. Arrows represent the direction of vibrotactile feedback activation. The color scale on the dots represents the intensity of vibration.}
 \label{fig:haptic_ill}
\end{figure}

Thus, the swarm is capable of simulating various surface textures through distributed vibration patterns and various frequency levels.

\subsection{User Safety}
\label{section:Safety}
For tactile communication, drones must be close enough to the user. The main danger, in this case, is the drone's propellers. That is why we paid attention to drones' path planning for avoiding collisions between the drones and the user's hand in our research. Positioning error (section \ref{subsection:experiment1}) does not exceed 0.11 m for the path planning method that we used for experiments. While there is a significant distance between the user's hands and the swarm, the tester's eyes are protected by a virtual reality headset. Thus, in the event of a system failure, the likelihood of causing minor harm to a person is decreased in comparison with large drones.

\subsection{Artificial Potential Field}
\label{section:APF}
To avoid internal swarm collisions at the start of haptic interaction, we applied a potential field method to build the route from the starting position of drones to the initial position of the user \cite{Li_2020}. 
Each drone is affected by two forces: the gravity force and the force of repulsion from obstacles. The force of gravity is directed from the drone to its position above the user's hand and is proportional to the distance between these points. The repulsion force acts on the drone only if the other drone or hand (obstacles) are in a sphere with a certain radius centered on the drone's position. The vectors directed from obstacles to the drone are multiplied by a coefficient inversely proportional to the square of the distance to obstacles and summed up.

\subsection{Swarm Impedance Control}
\label{section:impedance}

The position and velocity of each drone in the swarm are managed passively by the user. The drones follow the positions of the user's fingers to attach vibromotors and deliver vibrotactile feedback to the user's hand as it approaches objects in the VR environment. To facilitate a safe and natural behavior of the drone swarm, we implement an impedance control model utilized in \cite{Tsykunov_2018}. In this model, each drone's position is coupled to the user's hand and the positions of other drones in the swarm via virtual mass-spring-damper models (Fig. \ref{fig:impedance}). 

\begin{figure}[htb!]
\centering
 \includegraphics[width=0.85\linewidth]{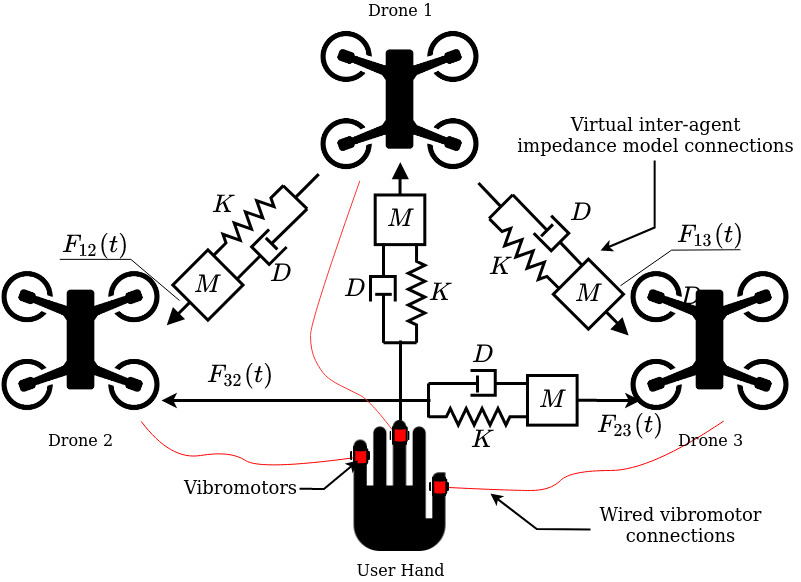}
 \caption{The topology and parameters of impedance links to achieve safe flight and compliant interaction.}
 \label{fig:impedance}
 \vspace{-0.2em}
\end{figure}

The configuration of these connections is determined by the specific application and desired swarm formation. The impedance provided by the virtual damped spring model prevents collisions within the swarm agents and with the user, thus providing a safe environment for object interaction.

Each virtual mass-spring-damper link dynamics is calculated with the position-based impedance control approach introduced in \cite{Hogan_1984} and can be represented by the second-order differential equation:

\vspace{-0.4em}

\begin{equation}
 \label{eq:1}
 M\Delta{\ddot x} + D\Delta{\dot x} + K\Delta x = F_\text{ext}(t) 
\end{equation}

\noindent
where $M$ is the virtual mass, $D$ is the damping coefficient of the virtual damper, $K$ is the stiffness of the virtual spring, $\Delta x$ is the difference between the current drone position $x_c$ and the desired position $x_d$, and $F_\text{ext}(t)$ is the externally applied force. For human-drone connections, the force is calculated using a human state parameter. In our implementation of the model, the external force for human-drone connections $F_\text{human}(t)$ is calculated as directly proportional to the user's hand velocity $v_\text{hand}$ in order to ensure a smooth trajectory for the drones with proper orientations and positions following the hand of the user, as defined in: 

\vspace{-0.4em}

\begin{equation}
 \label{eq:2}
 F_\text{human}(t) = K_v v_\text{hand}(t)
\end{equation}

\vspace{0.5em}

\noindent
where $K_v$ is the scaling coefficient, which can be selected to produce desirable feedback from the drones in response to the movement of the user's hand.
In order to solve the second-order differential equation (\ref{eq:1}), we utilize the solution discussed in \cite{Tsykunov_2018}, which rewrites the impedance equation as a state-space representation for discrete-time as follows:

\begin{equation}
 \label{eq:3}
 \begin{bmatrix}
 \Delta \dot x \\
 \Delta \ddot x
 \end{bmatrix} =
 A \begin{bmatrix}
 \Delta x \\
 \Delta \dot x
 \end{bmatrix} +
 BF_\text{ext}(t)
\end{equation}

\noindent
where, $A = \begin{bmatrix}
0 & 1 \\
-\frac{K}{M} & -\frac{D}{M}
\end{bmatrix}$ and $B = \begin{bmatrix}
0 \\
\frac{1}{M}
\end{bmatrix}$. The model is further simplified by integrating this equation in a discrete time-space, which is given by:

\vspace{-0.2em}

\begin{equation}
 \label{eq:4}
 \begin{bmatrix}
 \Delta x_{k+1} \\
 \Delta{\dot x}_{k+1}
 \end{bmatrix} =
 A_d \begin{bmatrix}
 \Delta x_k \\
 \Delta{\dot x}_{\text{k, VR headset}}
 \end{bmatrix} +
 B_d F_\text{ext}^k
\end{equation}

\vspace{0.5em}

\noindent
where $A_d$ and $B_d$ are defined using the matrix exponential and found using the Cayley-Hamilton theorem. The required calculations are defined in:

\begin{equation}
 \label{eq:5}
 A_d = e^{\lambda T} \begin{bmatrix}
 (1 - \lambda T) & T \\
 -bT & (1 - \lambda T - aT)
 \end{bmatrix}
\end{equation}

\vspace{0.7em}

\begin{equation}
 \label{eq:6}
 B_d = -\frac{c}{d} \begin{bmatrix}
 e^{\lambda T}(1 - \lambda T) - 1 \\
 -bTe^{\lambda T}
 \end{bmatrix}
\end{equation}

\vspace{0.5em}

\noindent
where $a = -\frac{D}{M}$, $b = -\frac{K}{M}$, $c = \frac{1}{M}$, and $\lambda$ is the eigenvalue of the matrix $A$. By selecting the parameters of the impedance model ($M$, $D$, $K$, and $K_v$) in such a way that the model is critically damped, the model is further simplified by ensuring both that $A$ only has one eigenvalue, $\lambda = \lambda_1 = \lambda_2$, and that solution is real. This eigenvalue can then be found as the root of the characteristic equation of the matrix $A$, given in:

\vspace{-0.4em}

\begin{equation}
 \label{eq:7}
 \lambda^2 + 2\zeta \omega_n \lambda + \omega_n^2 = 0
\end{equation}

\vspace{0.5em}

\noindent
where $\omega_n = \sqrt{\frac{K}{M}}$, $\zeta = \frac{D}{2\sqrt{MK}}$. By finding the total applied force $F_\text{ext}(t)$ for a given virtual link the target position and velocity of each drone can be calculated using Eq. (\ref{eq:4}). Each drone in the swarm then follows a certain position with given offset from the user's hand while avoiding collision with other drones in the swarm.

\section{Experimental Evaluation}
\label{subsection:experiment1}
\subsection{Impedance Parameters and Verification}

Both the impedance model parameters and the interlink topology configuration require additional investigation to ensure that the drone swarm can maintain a desirable formation while following the user's hand. Impedance model parameters were calculated to satisfy a critically damped response (i.e. satisfying $\zeta = 1$), in addition, the potential field method was tested. Three interlink network topologies were tested (Fig. \ref{fig:topology}).

\begin{figure}[htb!]
 \includegraphics[width=0.9\linewidth]{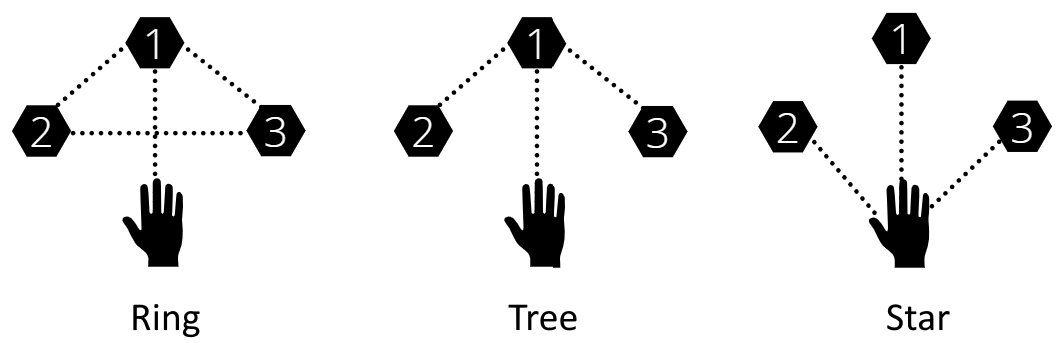}
 \caption{Evaluated impedance link topologies of drones in swarm for the impedance control model.}
 \label{fig:topology}
 \end{figure}

Several simulations were run to observe both the positions and velocities of the drones. These tests demonstrated that the most desirable set of parameters for the DandelionTouch formation was $M=1.9$, $D=12.6$, $K=20.88$, corresponding to a natural frequency $\omega_n = 3.3$. 

To test the real swarm behavior, we conducted an experiment with the drones following UR3 robotic arm that performed a square trajectory. The experiment was conducted 3 times for each formation, after which an average result was calculated. Fig. \ref{fig:position_plots} shows the trajectories of the swarm following the robot's end-effector.

\begin{figure}[htb!]
 \includegraphics[width=0.95\linewidth]{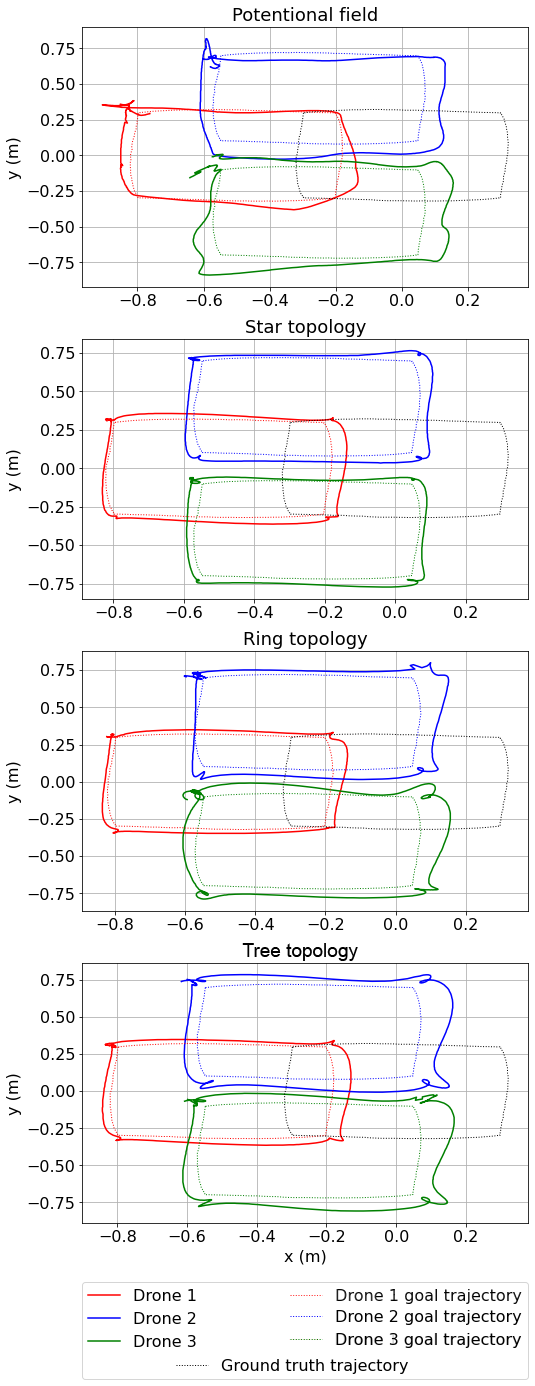}
 \caption{Trajectories of drones with different impedance link topologies for the three-agent swarm with following parameters $M=1.9$, $D=12.6$, $K=20.88$.}
 \label{fig:position_plots}
 \end{figure}

The experimental results showed that potential fields method is the less accurate comparing to the impedance control. The quantitative assessment is shown in Table \ref{tap:error}.

\begin{table}[h!]
\centering{
\caption{Positional error of drones with Impedance and Potential field swarm control algorithms.}
\setlength{\tabcolsep}{6pt} 
\renewcommand{\arraystretch}{1}
\begin{tabular}{ | c | c | c | c | c | }
\hline
\label{tap:error}
\begin{tabular}{p{1.5cm}cp{4cm}} Swarm control approach \end{tabular} & \multicolumn{2}{|c|}{Mean error, m} & \multicolumn{2}{|c|}{Max error, m} \\ \cline{2-3} \cline{4-5}
 & x & y & x & y \\ \hline
Impedance, Ring t.& 0.13 & 0.14 & 0.35 & 0.36 \\\hline
Impedance, Tree t. & 0.14 & 0.16 & 0.35 & 0.37 \\\hline
Impedance, Star t. & 0.10 & 0.11 & 0.27 & 0.29 \\\hline
Potential f. & 0.22 & 0.22 & 0.49 & 0.45 \\ 
\hline
\end{tabular}}
\end{table}

Using these results, the MSE for the position of drones (using the difference in x and y coordinates from the expected trajectory) was found to be $0.10$ m for x and $0.11$ m for y for the Star topology, which is the best configuration tested. The velocities of drones following a trajectory with the potential field method was significantly reduced to increase the stability of the drone flight (Table \ref{tap:speed}). The parameters of the Impedance Control method correspond to the speed of the robot. In Fig. \ref{fig:velocity_plot}, we observe a 1.3-second time delay for the impedance control.
 
 \begin{figure}[htb!]
 \includegraphics[width=0.9\linewidth]{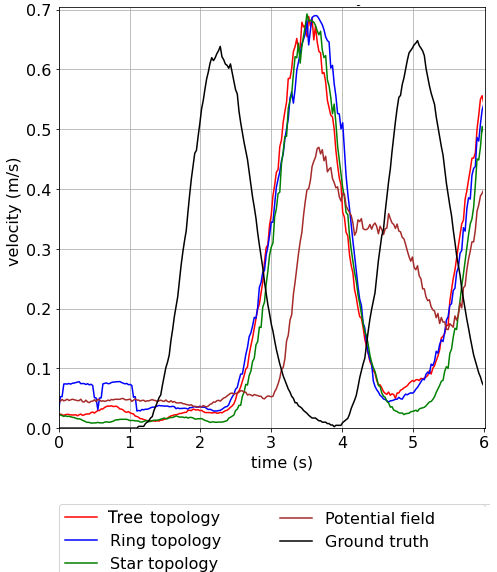}
 \caption{Mean velocities of the swarm drones in XY horizontal plane during testing of each topology.}
 \label{fig:velocity_plot}
 \vspace{-0.4em}
\end{figure}

The potential field method generates a more continuous motion with a lower acceleration value while significantly increasing the positioning error.
Thus, for the experiments, we have chosen the Impedance Control method for following the user's hand and the method of potential fields for approaching the initial positions. The smoother movement explains the latter choice towards a distant target.

\begin{table}[]
\centering{
\caption{Velocity of drones with Impedance and Potential field swarm control algorithms.}
\setlength{\tabcolsep}{6pt} 
\renewcommand{\arraystretch}{1}

\begin{tabular}{ | c | c | c | }
\hline
\label{tap:speed}
 Swarm control approach & Max velocity, m/s & Mean velocity, m/s\\ \hline
Impedance, Ring t.& 0.69 & 0.24  \\\hline
Impedance, Tree t. & 0.70 & 0.24 \\\hline
Impedance, Star t. & 0.69 & 0.22 \\\hline
Potential f. & 0.47 & 0.20  \\ \hline
Ground truth & 0.65 & 0.18  \\ \hline
\end{tabular}}
\end{table}

\section{User Study on Tactile Pattern Recognition}
We conducted a user evaluation of haptic interaction by ERM motors with patterns proposed in section \ref{section:haptic_patterns}. For this, we analyzed the recognition rate of three texture patterns with four possible directions of motion.

\subsubsection*{Participants}
We invited seven participants aged 22 to 28 years (mean = 24.7, std = 1.98) to test DandelionTouch system. Three of them have never interacted with haptic devices before, two used them several times, and two of the participants were familiar with CV-based systems or had some experience with gesture recognition. The participants were informed about the experiment and agreed to the consent form according to the laboratory's internal regulations.

\subsubsection*{Procedure} Before the study, the experimenter explained the purpose of the haptic interface to each participant and demonstrated the vibrotactile feedback patterns for each of the twelve combinations of 3 textures: rigid, elastic, and soft, and 4 motion direction: forward, backward, right, left (Table \ref{tap:patterns}). The demonstration was provided, at first, with additional visual feedback, and then at least one time blindly. The user was asked to wear an Oculus Quest headset and interact with the virtual object during the experiment (Fig.\ref{fig:ex1_setup}). 

\begin{table}[]
\centering{
\caption{Tactile patterns for representing the direction of motion and type of surface.}
\setlength{\tabcolsep}{6pt} 
\renewcommand{\arraystretch}{1}

\begin{tabular}{ | c | c | c | c | }
\hline
\label{tap:patterns}
   & Soft (3.3Hz) & Elastic (8Hz) & Rigid (100Hz)\\ \hline
Forward & SF & EF & RF \\\hline
Backward & SB & EB & RB \\\hline
Right & SR & ER & RR \\\hline
Left & SL & EL & RL  \\ \hline
\end{tabular}}
\end{table}

\begin{figure}[h!]
    \centering
    \includegraphics[width=1\linewidth]{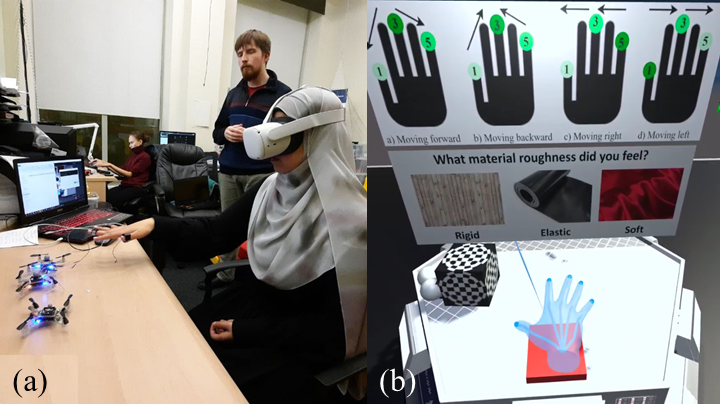}
    \caption{(a) The experimental setup of the human perception of various vibrotactile patterns. (b) The virtual environment with surface sample for pattern recognition.}
    \label{fig:ex1_setup}
\end{figure}

After receiving the haptic feedback from the drones via 3 fingers, the user informed instructors about the recognized pattern and then switched the pattern by pressing the virtual button with another hand. Each combination of motion direction and texture was presented 2 times blindly in random order. Thus, 24 patterns were provided in total to each participant in each evaluation.

\subsubsection*{Results}

In order to evaluate the statistical significance of the differences between patterns, we analyzed the results of the user study using single factor repeated-measures ANOVA, with a chosen significance level of p $<$ 0.05. According to the ANOVA results, there is a statistical significant difference in the recognition rates for the different patterns, $F (11,72) = 5.42240, p = 3\cdot10^{-4}$. 

\newcolumntype{A}{>{\raggedright}m{0.03\linewidth}}
\newcolumntype{g}{>{\columncolor{Gray}}c}
\newcolumntype{g}{>{\columncolor{Gray}}c}
\begin{table}[!ht]

\caption{Confusion matrix of shape pattern recognition}
\label{tab:conf_matrix}
\addtolength{\tabcolsep}{-4pt}
\begin{tabular}{|l|l|l|l|l|l|l|l|l|l|l|l|l|}
\hline

\multicolumn{1}{|c|}{} &
  \multicolumn{12}{c|}{\textit{Answers   (Predicted Class)}}\\ \cline{3-6} 
\hline
\textbf{Real} & SF & SB & SR & SL & EF & EB & ER & EL & RF & RB & RR & RL\\
\hline
SF & \cellcolor{Gray}\textbf{0.72} & \cellcolor{Gray2}\textbf{0.21} & 0.0 & \cellcolor{Gray2}\textbf{0.07} & 0.0 & 0.0& 0.0& 0.0 & 0.0 & 0.0 & 0.0 & 0.0\\\hline
SB & 0.0 &\cellcolor{Gray}\textbf{0.65} & \cellcolor{Gray2}\textbf{0.14} & \cellcolor{Gray2}\textbf{0.14} & 0.0 & 0.0& \cellcolor{Gray2}\textbf{0.07}& 0.0 & 0.0 & 0.0 & 0.0 & 0.0\\\hline
SR & \cellcolor{Gray2}\textbf{0.07} & \cellcolor{Gray2}\textbf{0.21} & \cellcolor{Gray}\textbf{0.72} & 0.0 & 0.0 & 0.0& 0.0& 0.0 & 0.0 & 0.0 & 0.0 & 0.0\\\hline
SL & \cellcolor{Gray2}\textbf{0.21} & \cellcolor{Gray2}\textbf{0.07} & \cellcolor{Gray2}\textbf{0.14} & \cellcolor{Gray}\textbf{0.58} & 0.0 & 0.0& 0.0& 0.0 & 0.0 & 0.0 & 0.0 & 0.0\\\hline
EF & \cellcolor{Gray2}\textbf{0.07} & 0.0 & 0.0 & 0.0 & \cellcolor{Gray}\textbf{0.65} & \cellcolor{Gray2}\textbf{0.21} & 0.0& 0.0 & \cellcolor{Gray2}\textbf{0.07} & 0.0 & 0.0 & 0.0\\\hline
EB & 0.0 & \cellcolor{Gray2}\textbf{0.07} & 0.0 & 0.0 & \cellcolor{Gray2}\textbf{0.07} & \cellcolor{Gray}\textbf{0.58}& \cellcolor{Gray2}\textbf{0.14} & 0.0 & 0.0 & \cellcolor{Gray2}\textbf{0.14} & 0.0 & 0.0\\\hline
ER & 0.0 & \cellcolor{Gray2}\textbf{0.07} & \cellcolor{Gray2}\textbf{0.21} & 0.0 & 0.0 & 0.0& \cellcolor{Gray}\textbf{0.72}& 0.0 & 0.0 & 0.0 & 0.0 & 0.0\\\hline
EL & \cellcolor{Gray2}\textbf{0.14} & 0.0 & 0.0 & 0.0 & \cellcolor{Gray2}\textbf{0.07} & \cellcolor{Gray2}\textbf{0.07}& 0.0& \cellcolor{Gray}\textbf{0.72} & 0.0 & 0.0 & 0.0 & 0.0\\\hline
RF & 0.0 & 0.0 & 0.0 & 0.0 & 0.0 & 0.0& 0.0& 0.0 & \cellcolor{Gray}\textbf{0.72} & \cellcolor{Gray2}\textbf{0.07} & \cellcolor{Gray2}\textbf{0.21} & 0.0\\\hline
RB & 0.0 & 0.0 & 0.0 & 0.0 & 0.0 & \cellcolor{Gray2}\textbf{0.07} & 0.0&\cellcolor{Gray2}\textbf{0.07} & 0.0 & \cellcolor{Gray}\textbf{0.79} & \cellcolor{Gray2}\textbf{0.07} & 0.0\\\hline
RR & 0.0 & 0.0 & 0.0 & 0.0 & 0.0 & 0.0& 0.0& 0.0 & 0.0 & \cellcolor{Gray2}\textbf{0.07} & \cellcolor{Gray}\textbf{0.93} & 0.0 \\\hline
RL & 0.0 & 0.0 & 0.0 & \cellcolor{Gray2}\textbf{0.07} & \cellcolor{Gray2}\textbf{0.07} & 0.0& 0.0& 0.0 & \cellcolor{Gray2}\textbf{0.14} & \cellcolor{Gray2}\textbf{0.07} & 0.0 & \cellcolor{Gray}\textbf{0.65}\\\hline
\end{tabular}
\end{table}

The paired t-tests with Welch’s test for unequal variances showed statistically significant differences between SB and SL patterns ($p=0.002 < 0.05$), ER and SB ($p=0.004 < 0.05$) and several other patterns. However, the results of paired t-tests between patterns SF and SB, SR and ER, SL and RL, RF and RR did not reveal any significant differences, so these patterns have nearly the same recognition rate. A two-way ANOVA showed statistically significant effect of both motion ($p=0.0021 < 0.05$) and material ($p=0.0008 < 0.05$) patterns on the recognition rate and significant interaction effect between these parameters: $F = 5.629, p= 7.8\cdot10^{-5} < 0.05$. The results revealed that in average in 70\% cases users were able to correctly recognize both the direction of movement of the hand and the type of material. Moreover, soft materials were recognized in 98$\%$ of cases, rigid in 93$\%$, and elastic in 80$\%$. Of the correct patterns, the most easily recognizable was the right direction of motion: 79$\%$. Finally, the subjects demonstrated higher confusion when estimating direction during vibration signals at 3.3 Hz frequency, suggesting the requirement for longer temporal delays between signals to each finger in future studies.

\section{Conclusion and Future work}

We have developed a novel haptic display DandelionTouch in which a human interacts with the virtual environment through vibrotactile feedback delivered to their fingers by a swarm of drones. The proposed interaction scenario utilizes the advantages of swarm mobility and distributed haptic feedback to achieve user interaction with soft virtual objects in a large area while eliminating the need for bulky haptic wearables. The impedance control over the swarm of drones was investigated, and the Star topology was evaluated to ensure stable hand tracking by drones (mean position error of 10.5 cm with 0.69 m/s maximal speed). A user study was conducted to investigate vibrotactile pattern recognition for different motion directions on different virtual surfaces. Soft materials were correctly recognized in 98\% of cases with the highest recognition of the right direction (79\%).  

DandelionTouch technology is most useful for scenarios where haptic feedback is beneficial or critical, but user exertion needs to be kept to a minimum.
One significant application is in the area of telemedicine to remotely treat non-surgical cases and carry out consultations where it is essential for the doctor to touch/feel the patient. For example, palpation is necessary to detect tumors and DandelionTouch would allow doctors to feel the patients remotely in situations such as pandemics.

\section{Acknowledgments}
The reported study was funded by RFBR and CNRS, project number 21-58-15006.

\bibliographystyle{IEEEtran}
\bibliography{bibliography}

\end{document}